\documentclass[10pt,twocolumn,letterpaper]{article}

\usepackage{cvpr}
\usepackage{times}
\usepackage{epsfig}
\usepackage{graphicx}
\usepackage{amsthm}
\usepackage{amsmath,amssymb}

\usepackage{url}
\usepackage{multirow}
\usepackage{booktabs}
\usepackage{algorithm2e}
\usepackage{mathtools}
\usepackage{tabularx}
\usepackage{varwidth}
\usepackage{stmaryrd}

\usepackage{fancyhdr}
\newcommand{\mbe}{\mathbf{e}}
\newcommand{\mbf}{\mathbf{f}}

\newcommand{\mbt}{\mathbf{t}}

\newcommand{\R}{\mathbb{R}}

\newcommand{\calE}{\mathcal{E}}
\newcommand{\calF}{\mathcal{F}}

\newcommand{\calN}{\mathcal{N}}

\newcommand{\calR}{\mathcal{R}}

\newcommand{\calX}{\mathcal{X}}

\newcommand{\inner}[1]{\left\langle#1\right\rangle}

\def\R{\mathbb{R}}

\def\calX{\mathcal{X}}

\newcommand{\ignore}[1]{}

\makeatletter
\DeclareRobustCommand\onedot{\futurelet\@let@token\@onedot}
\def\@onedot{\ifx\@let@token.\else.\null\fi\xspace}
\def\eg{{e.g}\onedot} 
\def\ie{{i.e}\onedot}

\makeatother

\expandafter\def\expandafter\normalsize\expandafter{%
\normalsize\setlength\abovedisplayskip{4pt}}

\expandafter\def\expandafter\normalsize\expandafter{%
\normalsize\setlength\belowdisplayskip{4pt}}

\usepackage[pagebackref=true,breaklinks=true,letterpaper=true,colorlinks,bookmarks=false]{hyperref}

\cvprfinalcopy 
\ifcvprfinal\fi

\begin{document}
\title{Semi-supervised Learning with Explicit Relationship Regularization}
\author{Kwang In Kim\\
Lancaster University\\
\and
James Tompkin\\
Harvard SEAS\\
\and
Hanspeter Pfister\\
Harvard SEAS
\and
Christian Theobalt\\
MPI for Informatics
}
\maketitle

\begin{abstract}
In many learning tasks, the structure of the target space of a function holds rich information about the relationships between evaluations of functions on different data points. Existing approaches attempt to exploit this relationship information implicitly by enforcing smoothness on function evaluations only. However, what happens if we explicitly regularize the relationships between function evaluations? Inspired by \emph{homophily}, we regularize based on a smooth \emph{relationship function}, either defined from the data or with labels. In experiments, we demonstrate that this significantly improves the performance of state-of-the-art algorithms in semi-supervised classification and in spectral data embedding for constrained clustering and dimensionality reduction.
\end{abstract}

\thispagestyle{fancy}

\section{Introduction}
Regularization attempts to prevent overfitting in ill-posed problems. It is commonly applied in semi-supervised learning tasks: Given a sparse labeling on $u$ data points with $s$ labels $\{(x_i,y_i)\}_{i=1}^{s}$, our goal is to learn a function $f$ which maps from an input space $M$ to a target space $N$. The lack of labels is compensated for by exploiting unlabeled data points to provide additional information, \eg, on the geometry of and/or probability distribution on $M$, from which the data are generated. Regularization tries to measure and limit the complexity of proposed $f$ solutions by preferring smaller training errors and placing restrictions on smoothness. This established approach helps solve a variety of learning problems, such as image and shape classification, tracking, and retrieval (\eg, \cite{ZhuGhaLaf03,ZhoBouLal04,EbeLarSch10,TanYuKim13}). 

The target space $N$ has a structure which may be defined implicitly or, in some applications, explicitly through pair-wise similarity or dissimilarity potentials. However, current regularization methods operate only on the function itself, and do not \emph{explicitly} consider the potentially rich informative structure of $N$ as something which can be used for regularization. Regularizing the structure --- or the \emph{relationships} --- is inspired by \emph{homophily}, which is actively used to predict relationships within social networks~\cite{McPSmiCo01,Agaetal09,KimTom12}: individuals with similar mutual friends, or local structure, are more likely to influence one another, e.g., if two individuals $A$ and $B$ are friends then they tend to have mutual friends, and if $A$ has an enemy $C$, then $B$ is also likely to be an enemy of $C$. We demonstrate that a priori knowledge of the smoothness of a relationship between entities can be exploited in inference on the entity itself.

\begin{figure}[t]
\centering
\includegraphics[width=0.85\linewidth]{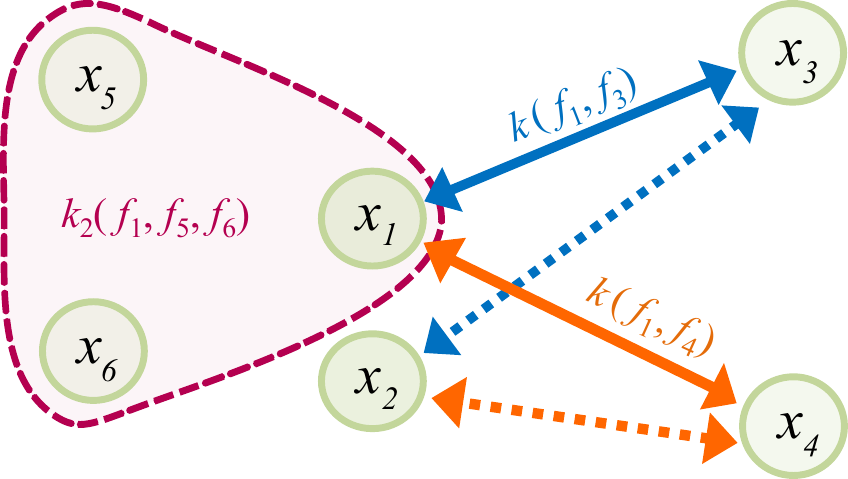}
\caption{\label{fig:diagram} If two data points $x_1$ and $x_2$ are close on the domain $M$ of $f$, then conventional regularizers enforce that the corresponding function values $f_1$ and $f_2$ in co-domain $N$ of $f$ are similar ($f_i\equiv f(x_i)$). We assume that relationships between pairs of function evaluations $f_i$ and $f_j$ are represented by smooth functions $k(f_i,f_j)$, \eg, a similarity measure. Our regularizer explicitly enforces that $k(f_1,f_j)$ and $k(f_2,f_j)$ are similar for any $j$. For instance, if $k(f_1,f_3)$ is large as $f_1$ and $f_3$ are similar, but $k(f_1,f_4)$ is small as $f_1$ and $f_4$ are dissimilar (solid arrows), then our algorithm enforces that $k(f_2,f_3)$ and $k(f_2,f_4)$ are large and small, respectively (dotted arrows), as $x_1$ and $x_2$ are close in $M$. The same principle applies to high-order relationships: if $k_2(f_1,f_5,f_6)$ represents a ternary relationship, \eg, a third-order correlation, the similarity of $k_2(f_1,f_5,f_6)$ and $k_2(f_2,f_5,f_6)$ is enforced.}
\end{figure}

One example that benefits from this principle occurs when \emph{relationship labels} are provided. In semi-supervised or \emph{constrained} spectral clustering~\cite{LiLiuTan09,RanHei12,WanDav10}, the labels are provided not on the underlying cluster assignment function $f$ but on the binary relationships $k$ between the function evaluations, as \emph{must-link} or \emph{cannot-link} labels. These are exploited by applying conventional regularization on $f$ with the condition that the constraints are satisfied. However, in this case, the relationship itself can also be a natural object to regularize (Fig.~\ref{fig:diagram}). Applying homophily, if $(x_1,x_3)$ \emph{must link}, \ie, if they belong to the same cluster, then a \emph{relationship function} $k$ on $N$ is defined such that $k(f(x_1),f(x_3))$ is positive. For point $x_2$, which is close to $x_1$ in $M$, we expect the relationship function $k(f(x_2),f(x_3))$ to be positive also.

In general, the relationship itself is not formally defined or observed; however, in many applications, certain relationships are manifested through a smooth function, where the number of arguments corresponds to the relationship degree, \eg, a distance metric is a function of two arguments. $k$ can be defined either directly from the data or from labels; either way, once the relationship is defined, regularization is independent of the existence of labels and therefore applies generally to any learning problem.

\subsection{Function-only and implicit relationships}
\label{sec:exploitingrelationships}
\label{sec:relatedwork}
We begin with a regularized empirical risk minimization framework where $f:M\to N$ minimizes the energy functional:
\begin{align}
\label{eq:obj}
\calE(f)= \sum_{i=1,\ldots,s} l(y_i, f(x_i)) + \lambda \calR(f),
\end{align}
where $\lambda$ is a regularization parameter, $\calR:N^{M}\to \R^+$ is the regularization functional that measures the \emph{complexity} of the input function, and $l:N\times N \to \R^+$ is the loss function. For simplicity, we assume that $N=\R^n$ and adopt the squared loss: $l(a,b)=\|a-b\|^2$, but our framework can be easily extended to other convex loss functions. Extension to non-Euclidean $N$ is also possible as discussed in Sec.~\ref{s:analyticregularization}.

While a variety of semi-supervised learning algorithms can potentially benefit from our approach (see \cite{ChaSchZie06} for a comprehensive survey), we focus on the successful class of graph Laplacian-based approaches. One of the best-established classes of regularizers is based on applying differential operators to $f$:
\begin{align}
\label{eq:genregenergy}
\calR_D(f) = \int_M \|[D f](x)\|^2 dV(x),
\end{align} 
where domain $M$ is the Riemannian manifold as is common in semi-supervised learning, and $dV(x)$ is the natural volume element of $M$. If $D$ is the first-order differential operator $\frac{d}{dx}$, then $\calR_D$ is the familiar \emph{harmonic energy} functional~\cite{BelNiy05,SteHeiSch10}:
\begin{align}
\label{eq:regenergy}
\calR^{\text{h}}(f) &= \int_M \|[\nabla f](x)\|_{T^*_x}^2 dV(x),
\end{align}
with Riemannian connection $\nabla$ in $M$, and cotangent space $T^*_x:=T^*_x(M)$ of $M$ at $x$~\cite{Lee97}. 

Roughly, this energy functional applies a differential operator to the input function and measures the corresponding squared norm. Minimizing this energy functional leads to a \emph{smooth} function with smaller first-order magnitudes. When $M$ is only indirectly observed through data point clouds, $\calR^{\text{h}}$ is instantiated based on the graph Laplacian~\cite{BelNiy05}, the performance of which has been demonstrated in numerous applications. 

Harmonic energy can be regarded as a first-order regularizer since it directly penalizes only variations of $f$. For relationships, denoted by double brackets, \eg, $\llbracket A,B\rrbracket$, this roughly corresponds to minimizing the pair-wise deviations between self-relationships $\llbracket f(x+\mathrm{d}x)\rrbracket$ and $\llbracket f(x)\rrbracket$, where $\llbracket A\rrbracket$ is simply as informative as $A$, with no consideration of relationships between entities.\footnote{A mathematically-precise relationship definition is obtained by equating the relationship with a set function $F: 2^M \to \R$. We do not adopt this definition since we focus on specific relationships instantiated through smooth kernels as defined in Sec.~\ref{s:relationdef}. In this sense, $\llbracket A\rrbracket$ can be identified with a set function defined on singletons, equivalent to a regular function on $M$.}

If we apply this first-order operator $\nabla$ twice to $f$, \ie, $D=\nabla^2$, we minimize the resulting second-order energy and penalize the deviations of the two pair-wise deviations $\llbracket f(x+\mathrm{d}x),f(x) \rrbracket$ and $\llbracket f(x-\mathrm{d}x),f(x) \rrbracket$. This can be regarded as an example of a second-order relationship regularizer, with the relationship defined as the difference between two entities. Higher-order relationship regularizers then enforce smoothness on relationships involving more than two entities by increasing the order of $D$. For instance, the state-of-the-art $p$-\emph{iterated Laplacian semi-norm}~\cite{ZhoBel11} measures smoothness of $(p-1)$-th order relationships.
\begin{align}
\label{eq:regilap}
\calR^{\text{p}}(f) &= \int_M f(x)[\Delta^p f](x)  dV(x).
\end{align}
However, existing differential operator-based regularizers focus only on \emph{local} relationships. By construction, $Df(x)$ is defined for an arbitrarily small open set containing $x$, and so it does not explicitly enforce smoothness over any pair $\llbracket f(x),f(x') \rrbracket$ and $\llbracket f(x''),f(x''') \rrbracket$ of relationships when all four input points $x$,$x'$,$x''$,$x'''$ do not lie within a small neighborhood --- even when $x$ and $x''$ are close. This property is shared by established regularizers in Euclidean space (\ie, $M$ is Euclidean): For instance, the well-known Gaussian kernel regularizer corresponds to Eq.~\ref{eq:genregenergy} with $D$ being a combination of powers of the Laplacian operator~\cite{Sch02}. 

\emph{Implicitly}, any existing regularization functional regularizes any high-order relationships, as smoothness on $f$ implies smoothness on pairs $\llbracket f(x),f(x') \rrbracket$. While apparently redundant, we will show experimentally that adding \emph{explicit} control over relationship regularization increases utility over existing function-only regularizers.

The success of local high-order derivative-based regularizers supports this claim: In 1D space, minimizing the first-order derivative norm as a regularizer implicitly minimizes all high-order derivative norms, as the only null space of the first-order derivative operator is the space of constant functions (as these have zero high-order derivatives). Nevertheless, the use of high-order derivative-based regularizers, \eg, thin plate spline and Gaussian regularizers, is strongly supported by their empirical performances. 

That high-order derivative-based regularizers can be considered as local high-order relationship regularizers, coupled with the success of these approaches over first-order (or non-relationship) regularizers, leads us to investigate the potential of `longer-range' relationship regularization. Among this various set of apparently-redundant regularizers, which leads to improved performance? We explore this potential and empirically validate that \emph{explicitly} exploiting rich structural information on non-local relationships improves existing regularization algorithms.

\section{Relationship regularization}

To begin, we focus on a specific class of relationships and discuss the ideal case where we know $M$ exactly. In Section \ref{sec:approximatingfrompointclouds}, we present a practical algorithm for when $M$ is indirectly represented as a sampled point cloud $\calX=\{x_1,\ldots,x_u\}$.

\subsection{Class of relationships}
\label{s:relationdef}

In many problems, $N$ has relationship structure that is either canonically specified by the problem or is given implicitly. In classification, the target space is the discrete space of class memberships. In this case, the natural relationship $\llbracket f(x),f(x') \rrbracket$ is binary: either \emph{same class} or \emph{different class}. In matching, $\llbracket f(x),f(x') \rrbracket$ is either \emph{match} or \emph{no match}.\footnote{$f$ may not be explicitly defined as the primary object in the relationship.} In Markov random fields (MRF), $N$ can be explicitly provided with a pair-wise potential $p:N\times N\to \R$, or an $n$-ary potential $q:N^n\to \R$~\cite{LaffMcCaPere01}. In many cases, these relationships represent similarity between pairs or $n$-tuples of entities; in general, any non-metric relationship can be defined, \eg, \emph{left of} or \emph{on top of} for generating topographic maps.

These relationships can be represented by an $n$-th order \emph{relationship function} $k$ defined on $N^n$, where $n$ is application specific. In principle, any relationship function can be regularized; for numerical optimization, we focus on $k$ that is \emph{smooth} wrt.~the input arguments (\ie, $k\in C^\infty(N^n)$). Specifically, for semi-supervised learning, we use a Gaussian relationship function $k$:
\begin{align}
\label{eq:gaussiansimilarity}
k(f(x),f(x'))=\exp\left(-\frac{(f(x)-f(x'))^2}{\sigma^2_f}\right)
\end{align}
where $\sigma^2_f>0$. We assume that $f\in C^\infty(M)$, which we regularize as aided by relationships. We obtain the final class membership $\{-1,1\}$ by thresholding the output space.

\subsection{Regularization on relations}
\label{s:analyticregularization}
Our proposed regularizer assumes the general cases where $N$ is a Riemannian manifold (though many examples, including our demonstrations, are Euclidean in $N$). First, we discuss a straightforward approach which is not computationally practical for large problems. Then, we develop this intuition further to arrive at a computationally-affordable solution.

We construct the regularizer of $f$ based on the regularization of relationship $k$ on the evaluations of $f$. First, we construct the \emph{pullback function}~\cite{Lee97} $f^*k$ of $k$ based on $f$:
\begin{align}
f^*k(x,x'):= k(f(x),f(x')).
\end{align}
This operation casts $k$, originally defined on $N^2$, into a function defined on $M^2$ so that it can be regularized based on the differential structure on $M^2$: Since $f^*k\in C^\infty (M^2)$, we can immediately extend the harmonic energy $\calR^{\text{h}}$ and the $p$-iterated Laplacian semi-norm $\calR^{\text{p}}$ as defined now on $M^2$ by noting that $f^*k$ can be regarded as a single-argument function on the product manifold $M^2$: 1) The tangent space for the point $(x,x')$ is defined based on the direct sum: $T_{(x,x')}:=T_{x}\oplus T_{x'}$; 2) The Riemannian metric is defined by $g_{M^2}(x_1+x_2,x_1'+x_2'):=g_{M}(x_1+x_2)+g_{M}(x_1'+x_2')$, which fixes the natural volume element $dV(x,x')$; 3) Based on 1) and 2), the differential structure $\nabla_{M^2}$ follows naturally from $\nabla_{M}$. 

The resulting new energy is in the same form as $\calR^{\text{h}}$ (Eq.~\ref{eq:regenergy}) except that its domain is now $M^2$ instead of $M$:
\begin{align}
\label{eq:prodenergy}
\calR^{\text{prod}}_k(f^*k) &= \int_{M^2} \|[\nabla f^*k](x,x')\|_{T^*_{(x,x')}}^2 dV(x,x').
\end{align}

The biggest obstacle to apply this straightforward construction to semi-supervised learning is its high computational complexity. When approximating $\calR^{\text{h}}$ and $\calR^{\text{p}}$ based on a sampled point cloud of size $u$, the corresponding approximations are calculated based on $u\times u$ matrices (Sec.~\ref{sec:approximatingfrompointclouds}). For the product manifold $M^2$, the approximations now require building regularization matrices of size $u^2\times u^2$, which become infeasible even for moderate $u$. 

Our approach is to make the roles of $x$ and $x'$ asymmetric in the regularization. For a given pair-wise relationship function $k$, we construct an auxiliary single-argument function $h$ and the corresponding pullback function $f^*h$ as:
\begin{align}
h_{y'}(y)&:=k(y,y')\in C^\infty (N),\\
f^*h_{x'}(x)&:=h_{f(x')}(f(x))\in C^\infty (M).
\end{align}

Now, we define new extensions of harmonic energy functional and $p$-th iterated Laplacian energy functional as:
\begin{align}
\label{eq:analylap}
\calR_{k}^{\text{h}}(f) &= \int_{M} \int_{M}\|\nabla f^*h_{x'}(x)\|_{T^*_x}^2dV(x) dV(x'),\\
\calR_{k}^{\text{p}}(f) &= \int_{M} \int_{M}  h_{x'}(x)[\Delta^p f^*h_{x'}(x)]dV(x) dV(x').
\label{eq:analyilap}
\end{align}

For each fixed $x'$ in the function, $f^*h_{x'}(x)$ encodes the relationship between $f(x)$ and $f(x')$, and since $f^*h_{x'}(x)$ is a function of a single variable $x\in M$, $\nabla f^*h_{x'}(x)$ lies in $T^*_x(M)$. This makes the interpretation of Eqs.~\ref{eq:analylap} and \ref{eq:analyilap} also straightforward: the inner integral measures the variation of $f^*h_{x'}(x)$ that corresponds to pair-wise relations between the fixed $x'$ and each value of $x$. In particular, when $k(a,b)$ measures the Euclidean distance between $a$ and $b$, the inner integral is zero only when the distances between each pair $\llbracket f(x),f(x') \rrbracket$ are identical for all $x\in M$. This does not require that $k$ is zero. Then, the outer integral averages $x'$ over the entire $M$. 

For an $n$-th order relationship function $q$, the corresponding $\calR_{q}$'s can be defined similarly through an $n$-times iterated integration: For each case, a pull-back function similar to $f^*h_{x'}(x)$ is defined as a $C^\infty$ function on $M$. An important advantage of this asymmetrization is that now the corresponding approximate regularization matrices retain the sizes of $u\times u$ (see Sec.~\ref{sec:approximatingfrompointclouds}) and accordingly they afford practical applications. 

It should also be noted that currently, our regularizer does not exploit the potential differential structure of the target manifold $N$. While the differential structure of $N$ is irrelevant in most applications we foresee, for interested readers, we note that in principle, our regularizer can take this structure into account by pulling it back to $M$, \ie, to use the \emph{pullback connection} $f^*\nabla^N$~\cite{SteHeiSch10}.

\subsection{Approximating $\calR_{k}$ from a sampled point cloud}
\label{sec:approximatingfrompointclouds}
In many practical applications, $M$ is not directly observed but indirectly represented as a sampled point cloud $\calX=\{x_1,\ldots,x_u\}$ and accordingly, we approximate $\calR_{k}$ based on evaluations of $f$ on $\calX$. For a given relationship function $k$, our approximate regularization functional to $\calR_{k}^{\text{h}}$ is defined as:
\begin{align}
\label{eq:discreg}
\widetilde{\calR_{k}^{\text{h}}}(\mbf) = tr[K^\top L K],
\end{align}
where $tr[\cdot]$ is the trace, $K_{ij}:=k(f(x_i),f(x_j))$, and $L (u\times u)$ is the graph Laplacian:
\begin{align}
\label{eq:graphLap}
L = D-W,
\end{align}
where $W_{ij}=\exp\left(-\frac{\|x_i-x_j\|}{\sigma_x^2}\right)$ when $x_i,x_j$ are $k$-nearest neighbors and $0$ when not, $\sigma_x^2$ is a hyper-parameter, and $D$ is a diagonal matrix containing the column sums of $W$. For exposition, we use the unnormalized graph Laplacian. However, our results straightforwardly extend to normalized graph Laplacian cases, which we use for all experiments (Sec.~\ref{s:complexity}).

By noting that the $i$-th column $K_{[:,i]}$ of $K$ corresponds to a discrete approximation of $f^*h_{x_i}(\cdot)$, the convergence of $\widetilde{\calR_{k}^{\text{h}}}$ to $\calR_{k}$ can be easily established based on the convergence results of the graph Laplacian to the Laplace-Beltrami operator~\cite{BelNiy05,HeiAudLix05}.

\par\noindent{\bf Proposition 1.}
\label{pr:prop}
Let $M$ be a connected, compact submanifold of $\R^M$ without boundary and $\calX_u=\{x_1,\ldots,x_u\}$ be sampled from a uniform distribution on $M$. Then, for $f\in C^\infty(M)$ and $k\in C^\infty(N\times N)$ and $\sigma^2_x(u)=u^{-\frac{1}{m+2+\alpha}}$ with $\alpha>0$,
\begin{align}
\label{e:prop}
\lim_{u\to\infty}\frac{\widetilde{\calR_{k}^{\text{h}}}(\mbf)}{u^3 (\sigma_x^2(u))^{m/2+1}} = \frac{\calR_{k}^{\text{h}}(f)}{V(M)^2},
\end{align}
in probability, where $V(M)$ is the volume of $M$.

\par\noindent{\bf Proof.}
The proof is similar to that of Theorem 4 by Zhou and Belkin \cite{ZhoBel11}. Since $f\in C^\infty(M)$ and $k\in C^\infty(N\times N)$, $f^*h_{x'}\in C^\infty(M)$. Then, applying the convergence result of graph Laplacian to $f^*h_{x_i}$ for a fixed $x_i$~\cite{BelNiy05}, we have $\forall x_j\in \calX$ in probability,
\begin{align}
\label{eq:convglap}
\lim_{u\to\infty} \frac{[L K_{[:,i]}]_{j}}{u (\sigma_x^2(u))^{m/2+1}} = \Delta f^*h_{x_i}(x_j).
\end{align}
For Eq.~\ref{e:prop}, we apply the law of large numbers and then Green's identity \cite{Lee97} for a compact manifold without boundary to Eq.~\ref{eq:convglap}:
\begin{align}
\label{e:green}
\hspace{1cm} \int_M f\Delta g dV(x)=-\int_M \inner{\nabla f, \nabla g}_{T_{x}^*}dV(x). \hspace{0.3cm} \Box
\end{align}

For simplicity, we assume a uniform sample distribution on $M$. However, this result extends to non-uniform underlying probability distributions $P$ on $M$ via Hein et al.~\cite{HeiAudLix05}. In this case, the integrand in Eq.~\ref{eq:analylap} is weighted by the corresponding density.

Similarly to $\calR_{k}^{\text{h}}$, the approximate regularization functional to $\calR_{k}^{\text{p}}$ is defined as:
\begin{align}
\widetilde{\calR_{k}^{\text{p}}}(\mbf) = tr[K^\top L^p K].
\end{align}
Given Prop.~\ref{pr:prop} conditions, the convergence of $\widetilde{\calR_{k}^{\text{p}}}$ to $\calR_{k}^{\text{p}}$ follows from Eq.~\ref{e:green} and the fact that $\Delta f \in C^\infty(M)$ for $f\in C^\infty(M)$.

\section{Semi-supervised learning}
\label{sec:semisupervisedlearning}
Given the two regularizers $\calR$ and $\calR_k$ (Eqs.~\ref{eq:regenergy} and \ref{eq:analylap} or Eqs.~\ref{eq:regilap} and \ref{eq:analyilap}) and the loss function ($l$; Eq.~\ref{eq:obj}), we state our semi-supervised learning algorithm:
\begin{align}
\label{eq:finalenergy}
\calE^k(\mbf) &= (\mbf-\mbt)^\top H(\mbf-\mbt) +\lambda_1\mbf^\top G \mbf+\lambda_2 tr[K^\top G K]\nonumber\\
&\approx \sum_{i=1,\ldots,s} l(y_i, f(x_i))+\lambda_1\calR^{\text{h}}(f)+\lambda_2 \calR_{k}^{\text{h}}(f),
\end{align}
where $\mbf=[f(x_1),\ldots,f(x_u)]^\top$, $H$ is a diagonal matrix, $H_{ii}=1$ if $i$-th data point is labeled ($0$ otherwise), $\lambda_1$ and $\lambda_2$ are regularization hyper-parameters, and $G$ is $L$ or $L^p$. For $\mbt$, if the $i$-th data point is labeled, $\mbt_{i}$ is the corresponding label $y_i$, or otherwise $0$.

While the first two summands in $\calE^k$ are convex with respect to $\mbf$, the third term is non-convex. We minimize $\calE^k$ based on conjugate gradient (CG) descent. We set the initial solution $\mbf^0$ as the minimizer of $\calE^k$ with $\lambda_2$ held fixed at 0, which can be analytically computed. Hence, the entire optimization process is deterministic.

With the Gaussian relationship function (Eq.~\ref{eq:gaussiansimilarity}), the gradient of each summand for the $t$-th function evaluation is:
\begin{align}
\label{e:dertrn}
\frac{\partial (\mbf-\mbt)^\top H(\mbf-\mbt)}{\partial \mbf}    \hspace{0.1cm} &= \hspace{0.1cm} 2H(\mbf-\mbt)\\
\label{e:derreg1}
\frac{\partial\mbf^\top G \mbf}{\partial \mbf}                  \hspace{0.1cm} &= \hspace{0.1cm} 2 G \mbf\\
\label{e:derreg2}
\frac{\partial tr[K^\top G K]}{\partial \mbf_t}                 \hspace{0.1cm} &= \hspace{0.1cm} 2 tr[K^\top G\frac{\partial K}{\partial \mbf_t}],
\end{align}
where $\mbf=[f(x_1),\ldots,f(x_u)]^\top$ and
\begin{align}
\label{e:derk}
\frac{\partial K_{ij}}{\partial \mbf_t} &= \left\{\begin{array}{ll}
-\frac{2(\mbf_i-\mbf_j)}{\sigma_f^2}K_{ij}& \text{if }i=t\\
-\frac{2(\mbf_j-\mbf_i)}{\sigma_f^2}K_{ij}& \text{else if }j=t.
\end{array}\right.
\end{align}

For (binary) classification problems, $y_i\in \{-1,1\}$. In Sec.~\ref{sec:relationshiplabels}, we discuss the dimensionality reduction problem where the output dimensionality $n$ is larger than $1$ and accordingly $f(x)$ is a vector.

\subsection{Sparsity}
\label{sec:sparsity}
Our empirical explicit relationship regularizer enforces smoothness across every possible pairwise evaluation of the function $f$. This leads to a dense matrix $K$ in Eq.~\ref{eq:finalenergy}. For large-scale problems, we can construct a sparse version of the regularizer by discarding the smoothness enforcement over the relationships that are evaluated for distant points, and focus only on local neighborhoods (not to be confused with the locality of the regularizer, \ie, neighborhood for graph Laplacian):
\begin{align}
\label{eq:sparseregl}
\calE_S^k(\mbf)=\lambda_2\sum_{i}\sum_{jk}(K_{ij}-K_{ik})^2 W_{jk}g_{ij}g_{ik},
\end{align}
where $g_{ij}=1$ if $x_i$ and $x_j$ are in a specified neighborhood $\calN_K$ and $g_{ij}=0$, otherwise. When the neighborhood size is infinite (\ie, $g=1$), $\calE_S^k$ is the same as the original regularizer in Eq.~\ref{eq:finalenergy}. Otherwise, $\calE_S^k$ enforces smoothness only for relationships that are defined for function evaluations of close input points.

\subsection{Relationship labels and spectral embedding}
\label{sec:relationshiplabels}
For some applications, the relationships $K$ themselves are natural variables of interest, and so training labels can be user provided. For instance, in spectral embedding such as for clustering and dimensionality reduction, \eg, in scientific visualization, where $f(x)\in\R^n$ with $n$ being the desired dimensionality, the absolute value of the function $f$ may be irrelevant while the relative \emph{spread} of the data are important. The user might provide expert rules to define which data points should be close to each other (\emph{must-link}) or not (\emph{cannot-link}). We can exploit this by penalizing the deviation of $K$ from the given relationship label $T$:
\begin{align}
\label{eq:relationlabel}
\calE_Q^k(\mbf)=\|(K-T).Q\|_\calF^2,
\end{align}
where $Q_{ij}=1$ if the label $T_{ij}$ is provided for a pair $(i,j)$, and $Q_{ij}=0$ otherwise. $T_{ij} = 1$ when $f(x_i)$ and $f(x_j)$ should be close to each other in the embedding space, and $T_{ij} = 0$ otherwise. $A.B$ is element-wise multiplication of two matrices $A$ and $B$, and $\|A\|_\calF$ is the Frobenius norm of $A$. In this case, our new energy functional is constructed as follows:
\begin{align}
\label{eq:dimredenergy}
\calE^k(\mbf) &= \|\mbf-\mbt\|^2 +\lambda_2 tr[K^\top G K]+\lambda_3\calE_Q^k(\mbf),
\end{align}
where we set the label $\mbt$ and the initial search solution $\mbf^0$ of the optimization as the results of standard spectral embedding obtained from a graph Laplacian-based algorithm: $\mbf^0=[\mbe_2,\ldots,\mbe_n]$ with $\mbe_i$ being the $i$-th eigenvector of $L$. Since each output $f(x)$ is a vector, our relationship function is adapted accordingly:
\begin{align}
\label{eq:vectorsimilarity}
k(f(x),f(x'))=\exp\left(-\frac{\|f(x)-f(x')\|^2}{\sigma^2_f}\right).
\end{align}
Minimizing Eq.~\ref{eq:relationlabel} over $\mbf$ is different from independently minimizing it for each output dimension since the outputs are tied across the dimensions through the relationship labels (Eq.~\ref{eq:dimredenergy}), and the regularizer (Eq.~\ref{eq:discreg}) is truly vector valued.

\section{Experiments}
\label{sec:experiments}

We compare the performance of our explicit relationship regularization (ERR, Eqs.~\ref{eq:analylap} and \ref{eq:analyilap}) by adapting two existing implicit relationship regularizations (IRR, Eqs.~\ref{eq:regenergy} and \ref{eq:regilap}): classic graph Laplacian~\cite{BelNiy05} and state-of-the-art iterated graph Laplacian~\cite{ZhoBel11}. To our knowledge, no algorithms exist which attempt to explicitly regularize relationships, even though they may implicitly attempt to do so (Sec.~\ref{sec:relatedwork}). The purpose of our experiments is to show the improvement that can come from explicit relationship regularization, using standard and state-of-the-art approaches as evidence. As such, we conducted a semi-supervised learning experiment for pattern classification with a set of standard machine learning databases. Code will be made available on the web.

\subsection{Semi-supervised classification}
\label{sec:patternclassification}
We use seven standard binary classification datasets for semi-supervised learning covering image digits (USPS), EEG signals (BCI), newsgroup categories (Text, Pcmac, Real-sim) and news reports (CCAT, GCAT) \cite{ChaSchZie06,ZhoBel11}. We randomly divide each dataset into three subsets: 50 labeled data points, 50 data points for validation for hyper-parameter selection, and the remaining unlabeled data points are used for evaluation. We average error rates for 10 experiments with different sets of labeled examples. To demonstrate sparsity for large datasets (Sec.~\ref{sec:sparsity}), we use the 60,000 point large MNIST dataset, with binary labels obtained in the same way as for the USPS dataset \cite{ChaSchZie06}. Here, $|\calN_K| = 200$, while the number of labeled and validation data points were fixed at 300 each. Due to the large size of the problem, the iterated graph Laplacian was not applicable for neither IRR nor ERR since taking the power of a sparse (Laplacian) matrix tends to produce a denser matrix.

Binary classification allows direct comparison of regularization performance and disregards multi-class combination method effects. However, to gain an insight into multi-class classification performance, we performed experiments with a 10-class dataset of 2,000 data points sampled from MNIST. For training and validation, we used 50 labels for each class. To facilitate representing the multi-class outputs, we learn a vector-valued function $f$ and the corresponding relationship function $k$ as defined in Eq.~\ref{eq:vectorsimilarity}.

For IRR, there are three parameters: $\sigma^2_x$, $k_N$, the $k$-nearest neighborhood size for the graph Laplacian construction (Eq.~\ref{eq:graphLap}), and regularization parameter $\lambda_1$. For ERR (Eq.~\ref{eq:analylap}), there are two more to be tuned: $\sigma^2_f$ for the Gaussian similarity relationship function $k$ (Eq.~\ref{eq:gaussiansimilarity}), and regularization parameters $\lambda_2$. We first find bounds for $\sigma^2_x$, $k_N$, and $\lambda_1$ around the optimal for IRR; then, we optimize $\sigma^2_f$ and $\lambda_2$ for ERR. This resulted in the total number of parameter evaluations for ERR being only slightly larger than that of IRR. For $\calR_{k}^{\text{p}}$ (Eq.~\ref{eq:analyilap}) there is an additional hyper-parameter $p$ that we fix at $2$ throughout the entire set of experiments.

\begin{table*}[t]
\centering
\caption{Classification performance as error rate for implicit and explicit relationship regularization (IRR and ERR), versus both graph Laplacian ($\widetilde{\calR_{k}^{\text{h}}}$) and iterated graph Laplacian ($\widetilde{\calR_{k}^{\text{p}}}$) regularizers, with added best-case parameters (BC; Sec.~\ref{sec:patternclassification}). Bold marks the best results. The performance improvement of ERR over IRR is calculated as the reduction of error rate (RER) in \%.}
\label{tab:clsresults}
\resizebox{0.8\linewidth}{!}
{
\begin{tabular}{ll rrrr rrrr p{1.66cm}}
\toprule
				&&USPS		&Text			&BCI 			&Pcmac		&Real-sim		&CCAT		&GCAT		&MNIST		&MNIST (multi-class)\\
\midrule
\multirow{7}{*}{\begin{varwidth}{6em} \centering Graph\\Laplacian\\$\widetilde{\calR_{k}^{\text{h}}}$\end{varwidth}\hspace{0.1cm}}
&IRR				&10.81		&43.13		&\textbf{42.98}	&14.97		&15.48		&26.08		&12.61 		&10.43		&\hfill 8.72    		\\
&ERR 				&\textbf{6.76}	&\textbf{35.13}	&43.38		&\textbf{11.62}	&\textbf{12.71}	&\textbf{25.92}	&\textbf{12.16} 	&\textbf{5.24}	&\hfill \textbf{7.03}    	\\
&RER (\%)			&37.46		&18.55		&-0.93		&22.38		&17.89		&0.06			&3.57 		&49.79		&\hfill 19.38    		\\
\cmidrule{2-11}
&IRR (BC)			&9.59			&37.91		&40.03		&13.61		&14.32		&20.80		&8.90 		&8.68			&\hfill 7.04    			\\
&ERR (BC)			&\textbf{4.44}	&\textbf{22.39}	&\textbf{38.95}	&\textbf{8.90}	&\textbf{10.23}	&\textbf{19.63}	&\textbf{8.39} 	&\textbf{4.90}	&\hfill \textbf{6.14}    	\\
&RER (\%)			&53.70		&40.94		&2.70			&34.61		&28.56		&5.63			&5.73 		&43.58		&\hfill 12.78    		\\
\midrule
\multirow{7}{*}{\begin{varwidth}{4em} \centering Iterated Graph Laplacian\\$\widetilde{\calR_{k}^{\text{p}}}$\end{varwidth}\hspace{0.1cm}}
&IRR				&4.80			&29.05		&\textbf{41.74}	&11.95		&12.36		&24.20		&10.97		& \multirow{6}{*}{\begin{varwidth}{3em} \centering N/A as matrix too dense\end{varwidth}\hspace{0.1cm}}		&\hfill 7.46    		\\
&ERR				&\textbf{3.71}	&\textbf{23.84}	&42.35		&\textbf{10.38}	&\textbf{11.52} 	&\textbf{21.31} 	&\textbf{9.48}	&			&\hfill  \textbf{6.74}   	\\
&RER (\%)			&22.71		&17.94		&-1.46		&13.14		&6.80			&11.94		&9.75			&			&\hfill  9.72   		\\
\cmidrule(l{2pt}r{2pt}){2-9}\cmidrule(l{45pt}r{2pt}){10-11}
&IRR	(BC)			&3.77			&24.40		&38.18		&10.07		&11.35		&18.94		&7.99			&			&\hfill 6.79    		\\
&ERR (BC)			&\textbf{2.33}	&\textbf{22.21}	&\textbf{37.58}	&\textbf{7.51}	&\textbf{9.68}	&\textbf{16.70}	&\textbf{7.26}	&			&\hfill  \textbf{6.14}    	\\
&RER (\%)			&38.20		&8.98			&1.57			&25.42		&14.71		&11.83		&9.14			&			&\hfill 9.65    		\\
\bottomrule
\end{tabular}
}
\end{table*}

\paragraph{Performance}
For all but one dataset, the error rate of ERR was lower than that of IRR when parameters were automatically chosen (Table~\ref{tab:clsresults}). This demonstrates the possible improvement of ERR over IRR and supports our claim that explicitly exploiting relationship information is useful. However, automatically optimizing the parameters with a limited number of labeled points can lead to overfitting (as observed in worse performance for ERR on BCI). Automatic tuning of hyper-parameters is still an open problem in semi-supervised learning where only a limited number of labeled examples are provided. 

We also report the performance of both algorithms when best-case (BC) hyper-parameters are provided (odd row blocks), and the performance difference between ERR and IRR is more pronounced. This indicates that ERR can potentially lead to larger improvements over IRR when the parameters are tuned properly (\eg, through user interaction). If the error rate surface with respect to the hyper-parameters is \emph{smooth}, then the user could decide the next search point based on the information gathered thus far. Our preliminary experiments showed that the error rate surface with respect to hyper-parameter \emph{is} smooth. Accordingly, the active sampling strategy can indeed be exercised (Table \ref{tab:clsresults}).

\subsection{Spectral embedding} 
\label{s:embedding}
Our algorithm is a general regularizer for Riemannian manifolds, and also supports explicit relationship labels. We use dimensionality reduction and clustering applications to show this with MNIST, full USPS, and standard UCI clustering datasets (Diabetes, Iris, Wine, Breast Cancer Wisconsin (BCW), and Pendigits). \emph{Must-link} and \emph{cannot-link} labels are based on ground truths for selected pairs. Note that relationship labels are \emph{weak} in that having a positive or negative label $T_{ij}$ for a pair $f_i$ and $f_j$ does not reveal the corresponding class information for either $y_i$ or $y_j$. 

\begin{figure}[t]
\centering
\includegraphics[width=0.7\linewidth]{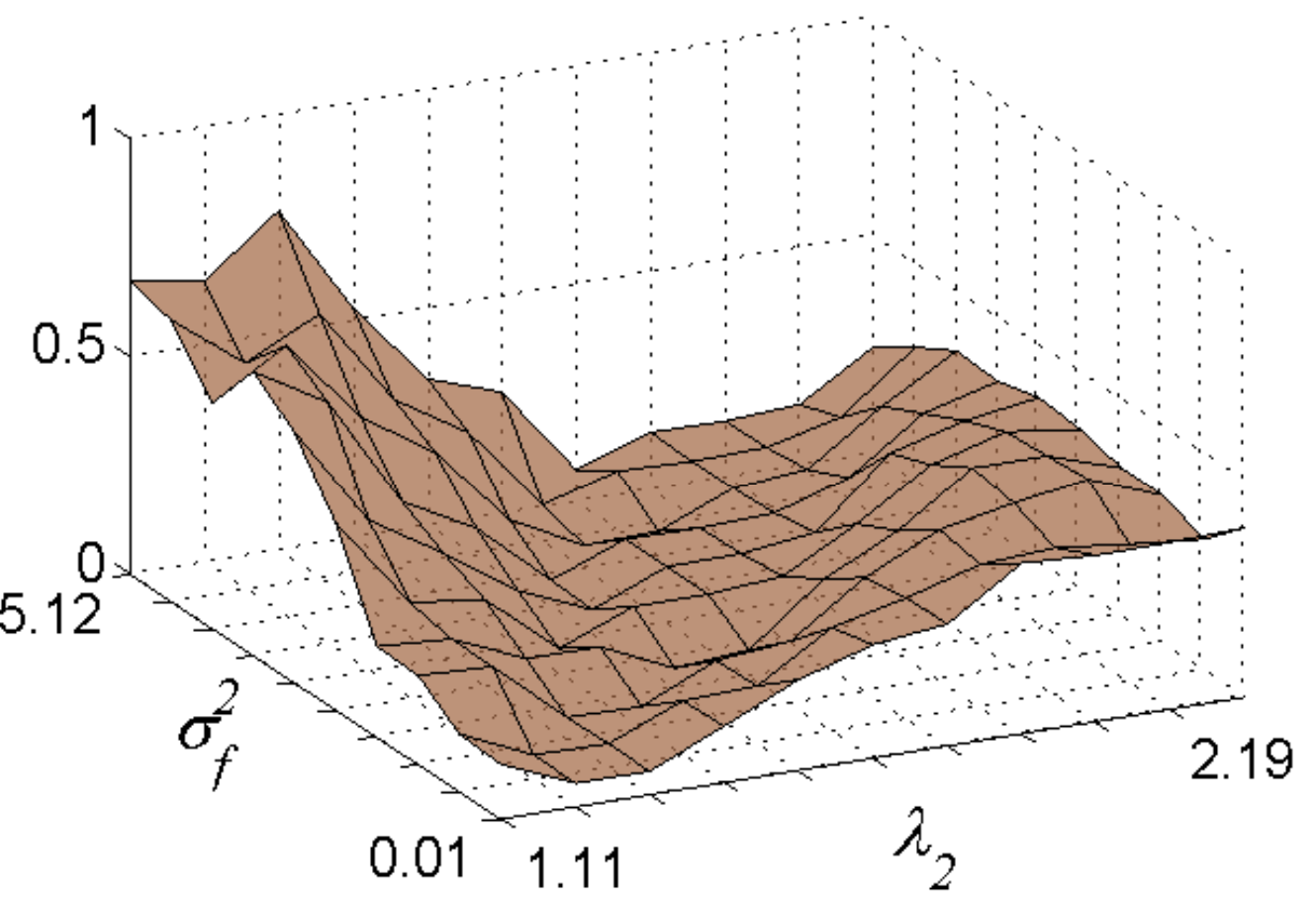}
\caption{\label{fig:hparam} Clustering performance (error rate) of the proposed algorithm on USPS dataset with hyper-parameters $\sigma_f^2$ and $\lambda_2$ ($\lambda_2'*S_R$) that vary in multiplicative intervals 2 and 3, respectively.}
\end{figure}

In general, for unsupervised learning such as clustering and dimensionality reduction, automatic tuning of hyper-parameters is infeasible as there is no ground-truth information. Following experimental convention \cite{BueHei09}, we set $k_N=10$ and $\sigma^2_x$ adaptively based on the average Euclidean distance of a point to its $k_N$ neighbors. In practice, the remaining hyper-parameters should be user tuned. To facilitate this process, we reduce the number of hyper-parameters to two, by first setting $\lambda_1=0$ (see Eq.~\ref{eq:dimredenergy}) and tying $\lambda_2$ and $\lambda_3$ by a new parameter $\lambda_2'$: We set the weight $\lambda_3$ of relationship labels at a relatively large value $10$ as these user labels should be regarded as quasi-hard constraints. The overall contribution of the $s_R$ relation labels is controlled by $\lambda_2'$, replacing $\lambda_2$ by $\lambda_2/s_R$. Figure~\ref{fig:hparam} shows that parameter tuning is feasible as performance varies smoothly with respect to the parameter space.

Again, while the hyper-parameters might be tuned based on user inspection in practice, to facilitate numerical evaluation for each dataset we randomly selected $s_R=250$ labels and optimized $\sigma^2_f$ and $\lambda_2'$ based on their respective ground-truth error measures (Sec.~\ref{sec:clustering}). These parameter values are fixed across \emph{all} $s_R$ values. For each value of $s_R$, we randomly sampled half the number of must-link and cannot-link labels, averaging error rates across 10 experiments. For comparison, we tuned the hyper-parameters of all competing algorithms (as described shortly) for each dataset and for each value of $s_R$, based on the ground-truth error rate, which is an advantage over our fixed parameters across $s_R$ values.

\subsubsection{Clustering}
\label{sec:clustering}
From the optimized $\mbf^*$, the final cluster label is assigned to each data point by applying $k$-means clustering on $\mbf^*$. Since $k$-means optimization is non-convex, we run it ten times with random initialization and choose the result that minimizes the \emph{normalized cut (NCut)} \cite{BueHei09} as it can be calculated without requiring any labels. We compare with the original spectral clustering, and three state-of-the-art algorithms which exploit explicit relationship labels: Constrained Clustering via Spectral Regularization (CCSR)~\cite{LiLiuTan09} and Flexible Constrained Spectral Clustering (CSP)~\cite{WanDav10} both optimize spectral energy ($\calR$) but under hard and soft constraints respectively (must-link and cannot-link), while Constrained 1-Spectral Clustering (COSC)~\cite{RanHei12} minimizes a continuous ($L1$) relaxation of the NCut under the same constraints. These algorithms significantly outperform existing (relationship-) constrained approaches, as well as unconstrained clustering algorithms~\cite{LiLiuTan09,RanHei12,WanDav10}. 

One major difference between those algorithms and ours is that they regularize $\mbf$ with constraints, while our algorithm explicitly regularizes relationships. We also compare with the more classical Spectral Learning algorithm (SL) that encodes the constraints into the weight matrix in building the graph Laplacian~\cite{KamKleMan03}. For CCSR and CSP, we used the code provided by the authors on their websites. Since CSP is designed for binary clustering, we only report the corresponding results of binary datasets (Diabetes, BCW). The clustering error is defined by summing the occurrences of errors for each cluster: a data points is counted as an error if its label is different from the dominant label of the cluster to which it belongs.

\begin{table*}[t]
\centering
\caption{Clustering performance as error rate for different constrained clustering algorithms.}
\label{tab:clustresults}
\resizebox{0.60\linewidth}{!}
{
\begin{tabular}{cl rrrr rrr}
\toprule
\# labels ($s_R$) & &Diabetes 			&BCW		&USPS		&MNIST			&Iris		&Wine		&Pendigits\\
\midrule
& Original &23.25 &33.02 &34.77 &13.05 &29.71 &34.96 &29.89 \\
\midrule
\multirow{5}{*}{\begin{varwidth}{6em} 50 \end{varwidth}\hspace{0.1cm}} 
& CSP  &\textbf{30.21} &3.25 &\multicolumn{5}{c}{N/A --- CSP is binary only}\\
& SL   &34.80 &34.99 &18.96 &30.72 &1.80 &32.64 &15.69 \\
& CCSR &30.99 &\textbf{2.75} &47.55 &59.20 &2.27 &29.49 &18.78 \\
& COSC &33.58 &9.59 &18.01 &24.04 &5.27 &36.57 &19.67 \\
& ERR  &33.50 &6.34 &\textbf{13.27} &\textbf{19.88} &\textbf{1.53} &\textbf{21.52} &\textbf{12.28} \\
\midrule
\multirow{5}{*}{\begin{varwidth}{6em} 100 \end{varwidth}\hspace{0.1cm}} 
& CSP  &31.08 &5.24 &\multicolumn{5}{c}{N/A --- CSP is binary only}\\
& SL   &34.01 &32.11 &18.11 &29.16 &1.47 &23.65 &14.23 \\
& CCSR &29.26 &\textbf{2.77} &37.78 &47.19 &2.07 &29.04 &17.41 \\
& COSC &32.15 &5.39 &18.32 &25.91 &1.67 &29.61 &13.75 \\
& ERR  &\textbf{27.85} &3.95 &\textbf{12.40} &\textbf{17.85} &\textbf{0.87} &\textbf{9.89} &\textbf{8.60} \\
\midrule
\multirow{5}{*}{\begin{varwidth}{6em} 250 \end{varwidth}\hspace{0.1cm}} 
& CSP  &29.91 &2.99 &\multicolumn{5}{c}{N/A --- CSP is binary only}\\
& SL   &28.26 &12.91 &\textbf{5.17} &17.39 &0.13 &2.42 &6.37 \\
& CCSR &29.05 &2.78 &20.84 &34.69 &2.00 &28.65 &13.52 \\
& COSC &12.38 &0.92 &18.12 &19.60 &0.13 &4.27 &3.13 \\
& ERR  &\textbf{12.36} &\textbf{0.64} &10.17 &\textbf{15.20} &\textbf{0.00} &\textbf{0.45} &\textbf{1.65} \\
\midrule
\multirow{5}{*}{\begin{varwidth}{6em} 500 \end{varwidth}\hspace{0.1cm}} 
& CSP  &28.19 &3.05 &\multicolumn{5}{c}{N/A --- CSP is binary only}\\
& SL   &17.77 &6.25 &8.24 &12.98 &\textbf{0.00} &\textbf{0.00} &5.81 \\
& CCSR &28.98 &2.87 &16.16 &28.86 &2.07 &27.87 &12.79 \\
& COSC &2.84 &\textbf{0.13} &17.30 &13.49 &\textbf{0.00} &0.06 &1.12 \\
& ERR  &\textbf{1.86} &0.15 &\textbf{5.14} &\textbf{12.83} &\textbf{0.00} &\textbf{0.00} &\textbf{1.09} \\
\midrule
\multirow{5}{*}{\begin{varwidth}{6em} 1,000 \end{varwidth}\hspace{0.1cm}} 
& CSP  &26.43 &2.80 &\multicolumn{5}{c}{N/A --- CSP is binary only}\\
& SL   &1.54 &0.44 &15.40 &24.67 &\textbf{0.00} &\textbf{0.00} &28.24 \\
& CCSR &29.34 &2.97 &11.69 &23.96 &1.93 &27.02 &12.29 \\
& COSC &0.39 &\textbf{0.00} &10.63 &9.79 &\textbf{0.00} &\textbf{0.00} &0.76 \\
& ERR  &\textbf{0.04} &\textbf{0.00} &\textbf{3.45} &\textbf{7.67} &\textbf{0.00} &\textbf{0.00} &\textbf{0.67} \\
\bottomrule
\end{tabular}
}
\end{table*}

\paragraph{Performance}
All algorithms that exploit relationship labels significantly improved over original spectral clustering (Table~\ref{tab:clustresults}). The CSP and CCSR were especially good for BCW when the number of labels $s_R$ is small. However, they failed to show steady performance increases as $s_R$ increases. Further, for Diabetes, both algorithms showed much higher error rates than other algorithms. On average, SL showed better performance over CSP and CCSR. However, for some datasets, it showed significant error rate increases when $s_R$ is too large, which shows application limitation. Overall, COSC and our algorithm (ERR) demonstrated steady decreases of error rates as $s_R$ increases. However, except for one case (BCW for $s_R=500$), our algorithm outperformed COSC by a large margin. For USPS, the error rates of COSC stayed high even when $s_R=1,000$: in the original spectral clustering result, multiple classes are merged into a single cluster, which leads to a single class dominating in multiple clusters. Classes 1 and 4 dominated in two clusters, respectively, and accordingly, classes 6 and 10 are absorbed. While ERR restored all classes when $s_R=500$, COSC failed even when $s_R=1,000$.

\subsubsection{Dimensionality reduction}
\label{sec:dimredexpr}
The target dimensionality $n$ was set at 2 for all experiments, \eg, for visualization applications, though any dimensionality is possible. We measured the error rate based on leave-one-out 1-nearest neighbor classification: For each point, we find its nearest neighbor and use the corresponding retrieved class label as the predicted label and measured the error rate. For comparison, we show the results of CCSR and SL. While both CCSR and SL were originally developed for clustering, they first perform spectral embedding to a given target dimension and then apply conventional clustering therein. Their embedding parts can be used for dimensionality reduction by choosing the target dimension accordingly.

\begin{table*}[t]
\centering
\caption{Leave-one-out classification performance as error rate for different dimensionality reduction algorithms.}
\label{tab:embedresults}
\resizebox{0.60\linewidth}{!}
{
\begin{tabular}{cl rrrr rrr}
\toprule
\# labels ($s_R$) & &Diabetes 			&BCW		&USPS		&MNIST			&Iris		&Wine		&Pendigits\\
\midrule
& Original &46.35 &9.37 &29.29 &34.48 &4.67 &28.09 & 15.92\\
\midrule
\multirow{3}{*}{\begin{varwidth}{6em} 50 \end{varwidth}\hspace{0.1cm}} 
& SL   &39.40 &6.50 &28.68 &33.29 &3.60 &31.35 &12.88 \\
& CCSR &36.59 &\textbf{3.91} &42.62 &33.70 &2.73 &34.55 &9.04 \\
& ERR  &\textbf{33.95} &4.77 &\textbf{5.34}  &\textbf{23.29} &\textbf{3.07} &\textbf{24.49} &\textbf{2.80} \\
\midrule
\multirow{3}{*}{\begin{varwidth}{6em} 100 \end{varwidth}\hspace{0.1cm}} 
& SL   &37.49 &7.13 &27.93 &33.15 &3.67 &30.28 &13.04 \\
& CCSR &37.21 &\textbf{4.04} &38.39 &34.98 &2.80 &33.60 &8.81 \\
& ERR  &\textbf{30.63} &4.10 &\textbf{5.35}  &\textbf{22.41} &\textbf{2.33} &\textbf{16.74} &\textbf{3.07} \\
\midrule
\multirow{3}{*}{\begin{varwidth}{6em} 250 \end{varwidth}\hspace{0.1cm}} 
& SL   &36.93 &7.10 &25.85 &31.05 &1.87 &20.34 &11.45 \\
& CCSR &37.38 &4.09 &29.24 &37.43 &3.33 &33.09 &8.92 \\
& ERR  &\textbf{24.92} &\textbf{3.41} &\textbf{5.30}  &\textbf{10.43} &\textbf{0.93} &\textbf{9.38}  &\textbf{2.60} \\
\midrule
\multirow{3}{*}{\begin{varwidth}{6em} 500 \end{varwidth}\hspace{0.1cm}} 
& SL   &24.88 &3.63 &22.48 &27.02 &0.60 &2.58  &10.82 \\
& CCSR &37.72 &4.07 &32.27 &46.42 &3.33 &31.57 &9.02 \\
& ERR  &\textbf{16.39} &\textbf{1.65} &\textbf{5.11}  &\textbf{6.62}  &\textbf{0.27} &\textbf{0.90}  &\textbf{2.58} \\
\midrule
\multirow{3}{*}{\begin{varwidth}{6em} 1,000 \end{varwidth}\hspace{0.1cm}} 
& SL   &11.78 &1.39 &17.25 &22.68 &\textbf{0.00} &0.11  &10.03 \\
& CCSR &38.06 &4.04 &36.93 &47.25 &3.33 &31.35 &9.53 \\
& ERR  &\textbf{9.53}  &\textbf{0.79} &\textbf{4.90}  &\textbf{6.31}  &\textbf{0.00} &\textbf{0.00}  &\textbf{2.20} \\
\bottomrule
\end{tabular}
}
\end{table*}

\begin{figure*}[t]
\centering
\includegraphics[width=0.9\linewidth,clip,trim=0mm 0mm 0mm 9.6mm]{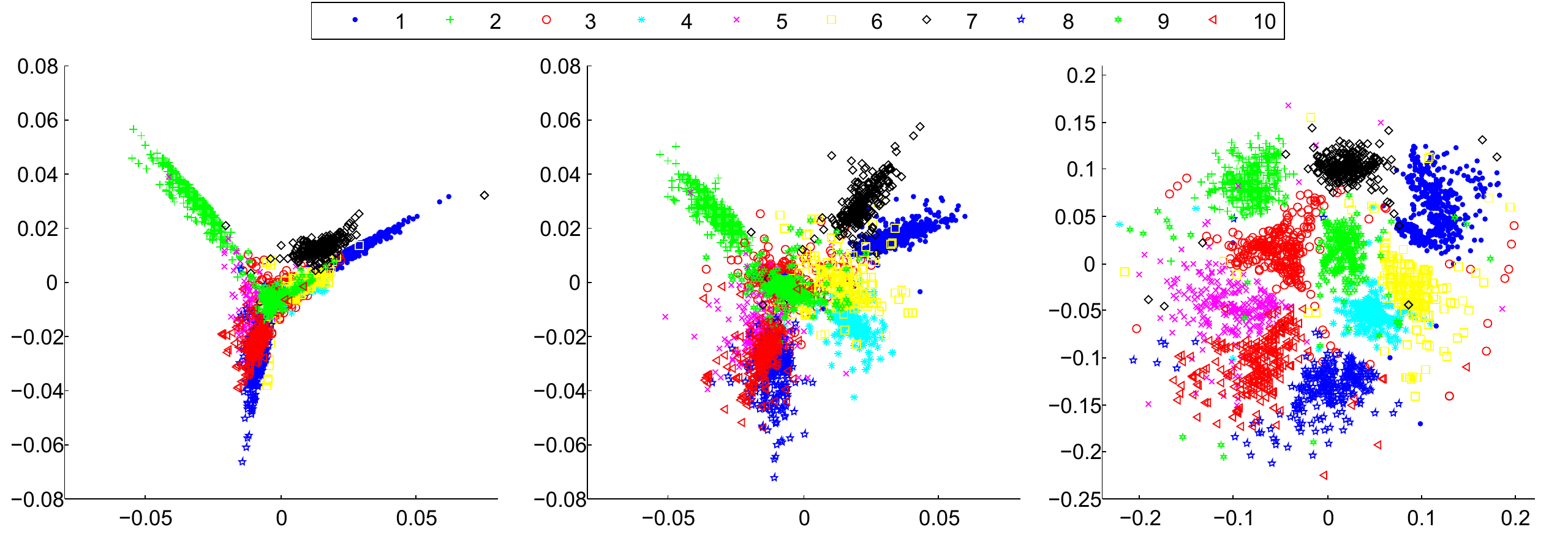}
\caption{\label{fig:embedexample} Embedding results for full 10-class USPS dataset ($s_R=100$); plots show only 2,000 data points for better visibility. \emph{Left:} Spectral embedding ($\mbt$ in Eq.~\ref{eq:dimredenergy}). \emph{Middle:} Minimizing 1) deviation from $\mbt$; 2) training error for relationship labels (Eq.~\ref{eq:relationlabel}), and 3) conventional graph Laplacian regularization energy ($\calE_M^k$ and $\calR$: Eqs.~\ref{eq:relationlabel} and \ref{eq:finalenergy} with $\lambda_2=0$). \emph{Right:} Our proposal ($\calE_M^k$ and $\calR_k$: Eq.~\ref{eq:dimredenergy}). Error rates (left to right): $28.30$, $27.53$, and $0.63$.}
\end{figure*}

\paragraph{Performance}
All algorithms improve over the original spectral dimensionality reduction (Table~\ref{tab:embedresults}), demonstrating the utility of relationship labels. CCSR was especially good for BCW, but it did not show noticeable improvement as $s_R$ increases. ERR and SL both showed steady error rate decreases while ERR significantly outperformed SL, demonstrating the utility of explicit relationship regularization. Figure~\ref{fig:embedexample} shows an example embedding.

\subsection{Complexity} 
\label{s:complexity}
For all experiments, following conventions, the graph Laplacians are normalized. We set the number of conjugate gradient (CG) steps to 50. This provides a moderate trade-off between the performance and accuracy: While we observed a steady increase in accuracy as the number of CG steps increased for pattern classification experiments, the rate of increase dropped significantly past 50. As indicated by the form of the energy functional (Eq.~\ref{eq:discreg}), when sparsity in relationships is not enforced (see Eq.~\ref{eq:sparseregl}), the time complexity of each gradient step is cubic in the number of data points. For pattern classification experiments with the USPS dataset (with 1500 data points), it took approximately 1.6 seconds for 50 CG step on NVIDIA GeForce 680 GPU, and 25 seconds on Intel Xeon 3.6GHz CPU; while the IRR took approximately 0.3 seconds on the same CPU: IRR can be solved analytically, while ERR must be solved iteratively.

\begin{table}[t]
\caption{Performance vs.~sparsity ($|\calN_K|$) for MNIST subsets ($s=100,u=2,000$). GPU optimization negates the need for sparsity for these problem sizes.}
\label{tab:sparsitylevel}
\resizebox{\linewidth}{!}
{
\centering
\begin{tabular}{c cccc cc}
\toprule
$|\calN_K|$         & 25    & 50    & 100   & 200   & full ERR  & IRR   \\
\midrule
Error ($\%$)        & 9.76  & 9.08  & 8.64  & 8.02  & 7.82      & 10.10 \\
Time CPU (sec.)     & 3     & 10    & 21    & 38    & 35        & 1     \\
Time GPU (sec.)     & \multicolumn{4}{c}{-}         & 3         & -     \\
\bottomrule
\end{tabular}
}
\end{table}

\subsection{Sparsity}
To gain an insight into the sparsity/performance trade-off, we performed experiments on a small subset ($u=2,000$) of the MNIST dataset such that direct performance comparison with dense regularization is possible (Table~\ref{tab:sparsitylevel}). Performance degrades gracefully as $|\calN_K|$ decreases. For this small dataset, the processing time of the sparse system when $|\calN_K|=200$ is longer than the full ERR due to the sparsification overhead. However, the complexity grows roughly linearly with respect to $u$, and thus sparsity makes ERR applicable to large-scale datasets. In Table~\ref{tab:clsresults}, we show the results of the full MNIST dataset with $|\calN_K|=200$.

\section{Discussion}
\label{sec:discussion}

We have only evaluated the binary relationship function $k$ with the single parameter $\sigma^2_f$, and different potential relationship function types could be explored. Further, we have only investigated binary relationship functions, and $n$-ary relationship functions are possible. In this case, the $K$ matrix in Eq.~\ref{eq:finalenergy} is replaced by a tensor, and the problem complexity increases, though it may still be possible to handle these cases by enforcing sparsity (Sec.~\ref{sec:sparsity}).

For the specific case of binary relationship functions regularized by the graph Laplacian (which corresponds to pair-wise regularization), our regularization energy functional (Eq.~\ref{eq:sparseregl}) can be regarded as a construction of a ternary relationship function: One can define a ternary clique as a summand of Eq.~\ref{eq:sparseregl}:
\begin{align}
q(f_i,f_j,f_k)=(K_{ij}-K_{ik})^2W_{jk}g_{ij}g_{ik}.
\end{align}
In this way, our algorithm can be viewed as a special case of an MRF. While, in general, the optimization with a ternary relationship function is computationally very demanding, the asymmetric roles of three arguments in our clique (see the last paragraph of Sec.~\ref{s:analyticregularization}) leads to a computationally affordable algorithm. In this respect, one of our main contributions is a method to construct a high-order clique from low-order cliques and the corresponding practical algorithm for semi-supervised learning.

In our semi-supervised learning experiments, we chose hyper-parameters based on separate validation sets. Heuristics can help set some hyper-parameters, e.g., for spectral embedding, we set $\sigma^2_x$ based on the average Euclidean distance of a point to its $k_N$ neighbors (Sec.~\ref{s:embedding}). For USPS, the corresponding average clustering error rate was around 20\% higher than when varying and manually selecting $\sigma^2_x$. This suggests that the heuristic can trade accuracy with hyper-parameter optimization time.

\section{Conclusion}
We have investigated \emph{explicit relationship regularization}, which, in addition to regularizing the function in semi-supervised learning, now regularizes the relationships between function evaluations through smooth relationship functions. This approach improves performance by a large margin in semi-supervised classification and in constrained spectral clustering applications, and facilitates a related algorithm in semi-supervised dimensionality reduction. We believe semi-supervised learning and constrained clustering algorithms will increase in importance in vision, \eg, recent works in pose estimation \cite{TanYuKim13}, and video segmentation \cite{KhoGalHei14}. Future work should consider what role our explicit relationship regularization plays on the effect of the statistical model, \eg, error bound.

\section*{Acknowledgements}
Kwang In Kim thanks EPSRC EP/M00533X/1 and EP/M006255/1, James Tompkin and Hanspeter Pfister thank NSF CGV-1110955, and James Tompkin and Christian Theobalt thank the Intel Visual Computing Institute.

\raggedbottom

\bibliographystyle{ieee}
\bibliography{./biblio}

\end{document}